\documentclass{article}

\usepackage[final, nonatbib]{nips_2018}

\usepackage{amsmath,amsfonts,amssymb,amsthm}
\usepackage{newpxtext} 
\usepackage{bbm}
\usepackage[utf8]{inputenc} 
\usepackage[T1]{fontenc}    
\usepackage{hyperref}       
\usepackage{url}            
\usepackage{booktabs}       
\usepackage{amsfonts}       
\usepackage{amsmath}       
\usepackage{nicefrac}       
\usepackage{microtype}      
\usepackage{graphicx}
\newcommand{\cmmnt}[1]{}

\usepackage[sorting=none]{biblatex}

\title{Interpretable Clustering via \\ Optimal Trees}

\author{
  Prof. Dimitris Bertsimas \\
  Sloan School of Management and Operations Research Center\\
  Massachusetts Institute of Technology\\
  Cambridge, MA 02142 \\
  \texttt{dbertsim@mit.edu} \\
  \And
  Agni Orfanoudaki \\
  Sloan School of Management and Operations Research Center\\
  Massachusetts Institute of Technology\\
  Cambridge, MA 02142 \\
  \texttt{agniorf@mit.edu} \\
  \AND
  Holly Wiberg \\ 
  Sloan School of Management and Operations Research Center\\
  Massachusetts Institute of Technology\\
  Cambridge, MA 02142 \\
  \texttt{hwiberg@mit.edu.} 
}

\begin{document}

\maketitle

\begin{abstract}
  State-of-the-art clustering algorithms use heuristics to partition the feature space and provide little insight into the rationale for cluster membership, limiting their interpretability. In healthcare applications, the latter poses a barrier to the adoption of these methods since medical researchers are required to provide detailed explanations of their decisions in order to gain patient trust and limit liability. We present a new unsupervised learning algorithm that leverages Mixed Integer Optimization techniques to generate interpretable \cmmnt{and globally optimal}tree-based clustering models. Utilizing the flexible framework of Optimal Trees \cite{bertsimas2017optimal}, our method approximates the globally optimal solution leading to high quality partitions of the feature space. Our algorithm, can incorporate various internal validation metrics, naturally determines the optimal number of clusters, and is able to account for mixed numeric and categorical data. It achieves comparable or superior performance on both synthetic and real world datasets when compared to K-Means while offering significantly higher interpretability.
\end{abstract}

\section{Introduction}

In the era of Electronic Medical Records (EMR) and advanced health monitoring, the huge amount of data generated is too complex and voluminous to be analyzed by traditional methods \cite{dataComplex}. Unsupervised learning methods are able to transform this heterogeneous data into meaningful information for decision making \cite{10.1007/11875604_54}. However, if the distinguishing characteristics of clusters are not easily identifiable, the results have limited utility. Interpretability is particularly important in a medical setting, where decision making can significantly impact individuals' disease trajectories.

There has been limited success in addressing the issue of cluster interpretability. The most popular method is the representation of a cluster of points by their centroid or by a set of distant points in the cluster which has been popular across various applications \cite{Radev2004}. This works well when the clusters are compact or isotropic but fails when the clusters are elongated or non-isotropic. Another common approach is the visualization of clusters in a two-dimensional graph using principle component analysis (PCA) projections \cite{jolliffe2011,rao1964}. However, in reducing the dimensionality of the feature space, PCA obscures the relationship between the clusters and the original variables.

Tree-based supervised learning methods such as CART \cite{breiman1984classification} and Optimal Classification Trees \cite{bertsimas2017optimal} are a natural fit for problems that prioritize interpretability. This is a two-step process: a traditional clustering methods give cluster assignments which can be used as class labels. The data can then be fit using a supervised classification tree; the decision paths leading to each cluster's leaves give insight into the differentiating features \cite{Hancock2003}. These trees give an explicit delineation of cluster attributes, but the two-step process does not prioritize interpretibility during cluster creation.

Motivated by the limitations of existing solutions to interpretable clustering, we propose a new tree-based machine learning algorithm, where interpretability is taken into consideration during cluster creation rather than considered as a later analysis step. Our method, called Interpretable Clustering via Optimal Trees (ICOT), builds upon the algorithm of Optimal Classification Trees \cite{bd17} and extends it to the unsupervised setting. Our algorithm constructs a tree with a perspective of global optimality rather than taking a greedy approach. It is formulated as a mixed-integer optimization problem which can be solved using an iterative coordinate-descent approach that scales to larger problems, approximating the globally optimal solution. 

We propose an unsupervised learning algorithm that solves the task at hand using an optimization lens while providing the user with more accurate and interpretable results based on the feature vectors. We use well-established validation criteria, such as the Silhouette Metric \cite{Rousseeuw1987} and Dunn Index \cite{Dunn1974}, as the algorithm's objective function taking into account both the inner-cluster density as well as the intra-cluster separation. Our technique renders the tuning of the number of clusters redundant, making it easy to be used by medical researchers. We test the performance of our method compared to K-Means in available datasets from the Fundamental Clustering Problems Suite (FCPS) and a real-world example from the Framingham Heart Study. We demonstrate its superior performance in data with different levels of variance and compactness as well as its ability to provide us with interpretable clusters. 

\section{From Supervised to Unsupervised Learning}

ICOT builds trees through a modification of Optimal Classification Trees (OCT), a globally optimal tree-based algorithm \cite{bertsimas2017optimal}. The resultant tree provides an explicit characterization of membership in a cluster, represented by a single leaf, through the path of feature splits. ICOT is formulated as a mixed-integer optimization problem (see Supplementary Material), which creates a decision tree that minimizes a chosen loss function and assigns each observation to a leaf based on parallel feature splits. The algorithm is implemented using a coordinate-descent procedure which allows it to scale to much higher dimensions than directly solving the mixed-integer formulation, well-approximating the optimal solution while still abiding by the same core principles.

ICOT initializes a greedy tree and then runs a local search procedure until the objective value, a cluster quality measure, converges. This process is repeated from many different starting greedy trees, generating many candidate clustering trees. The final tree is chosen as the one with the highest cluster quality score across all candidate trees. This single tree is returned as the output to the algorithm. 

To form the initial greedy tree, we start with a single node and scan over potential splits on a randomly chosen feature. For each potential threshold for splitting observations into the lower and upper leaves, we compute the global score for the resultant assignment of the proposed split. After scanning through all thresholds, we choose the one that gives highest score and update the node to add the split if this score exceeds the global score of the current assignment. We perform the same search for each leaf that gets added to the tree, continuing until either the maximum tree depth is reached or no further improvement in our objective value is achieved through splitting a leaf. 

Following the creation of the greedy tree, we begin the local search procedure. Nodes are visited in a randomly chosen order, and various modifications are considered. A "split" node (i.e. a node that is not a leaf) can be deleted, in which case it is replaced with either its lower or upper subtree, or a new split can be made at the node using a different feature and threshold. A leaf node can be further split and thus create two leaves. At each node, the algorithm finds the best possible change and then makes the proposed change only if it improves the objective from its current value. The algorithm terminates when the objective value converges.

\subsection{Model Parameters}
There are several user-defined inputs to the algorithm that give the user flexibility in their evaluation criterion and tree depth.

\subsubsection{Cluster Quality Measure}
The chosen loss function must consider the global assignment of observations to clusters. The score of a clustering assignment depends on both the \textbf{compactness} of the observations within a single cluster, as well as its \textbf{separation} from observations in other clusters. Several internal validation metrics have been proposed to balance these two objectives \cite{liu2010}. Two common criteria, the Silhouette and Dunn Index scores, are outlined below.

\paragraph{Silhouette Metric} The Silhouette Metric introduced by \cite{Rousseeuw1987} compares the distance from an observation to other observations in its cluster relative to the distance from the observation to other observations in the second closest cluster. The silhouette metric for observation $i$ is computed as follows: 
\begin{equation}
s(i) = \frac{b(i) - a(i)}{\max(b(i), a(i))}
\end{equation}
where $a(i)$ is the average distance from observation $i$ to the other points in its cluster, and $b(i)$ is the average distance from observation $i$ to the points in the second closest cluster (i.e. $\min_k b(i,k)$ where $b(i,k)$ is the average distance of $i$ to points in cluster $k$, minimized over all clusters $k$ other than the cluster that point $i$ is assigned to). This score ranges from -1 to 1, where a higher score is better. These individual scores can be averaged to reflect the quality of the global assignment.

\paragraph{Dunn Index}
The Dunn Index \cite{Dunn1974} characterizes compactness as the maximum distance between observations in the same cluster, and separation as the minimum distance between two observations in different clusters. The metric is computed as the ratio of the  minimum inter-cluster separation to the maximum intra-cluster distance. A high score is better, since it signifies that the distance between clusters is large relative to the distance between points within a cluster.

\subsubsection{Tree Depth and Complexity}
The natural balance between separation and compactness in the ICOT splitting process allows for a more exploratory approach to clustering. It eliminates the need for setting an explicit K parameter, which is typically required in both partitional and hierarchical clustering methods. The tree continues to split until further splits no longer improve the quality of the overall assignment, and so the final number of leaves represent the optimal number of clusters. The maximum depth can be used to impose an upper bound on a reasonable number of clusters if desired. 

\section{Results}

\subsection{Synthetic Datasets}
We evaluated ICOT on the Fundamental Clustering Problems Suite datasets (FCPS) ~\cite{FCPS}, a standard set of synthetic datasets for unsupervised learning evaluation. We compared three methods: ICOT, K-Means, and Optimal Classification Trees (OCT), in which we use K-Means clusters as class labels and approach the clustering tree creation as a supervised classification problem. These methods serve as benchmarks: K-Means is the current standard clustering practice, and OCT represents a method of building interpretable clustering trees as a two-step process using existing methods. Table~\ref{tab:results} shows the comparison of these methods along with the true FCPS labels, evaluated with both the Silhouette Metric and Dunn Index. The asterisks indicate the best score across all methods for each criterion.

\begin{table}[]
\begin{tabular}{lllllllll}
 & \multicolumn{4}{c}{\textbf{Silhouette Metric}} & \multicolumn{4}{c}{\textbf{Dunn Index}} \\ \hline
\textbf{Dataset} & \textbf{ICOT} & \textbf{K-Means} & \textbf{OCT} & \textbf{Truth} & \textbf{ICOT} & \textbf{K-Means} & \textbf{OCT} & \textbf{Truth} \\ \hline
\multicolumn{1}{l|}{Atom} & 0.491 & 0.601* & 0.441 & \multicolumn{1}{l|}{0.311} & 0.137 & 0.029 & 0.027 & 0.371* \\
\multicolumn{1}{l|}{Chainlink} & 0.395 & 0.405* & 0.282 & \multicolumn{1}{l|}{0.158} & 0.031 & 0.018 & 0.017 & 0.265* \\
\multicolumn{1}{l|}{EngyTime} & 0.573* & 0.438 & 0.410 & \multicolumn{1}{l|}{0.398} & 0.064* & 0.003 & 0.001 & 0.000 \\
\multicolumn{1}{l|}{Hepta} & 0.455 & 0.702* & 0.373 & \multicolumn{1}{l|}{0.702*} & 0.357 & 1.076* & 0.027 & 1.076* \\
\multicolumn{1}{l|}{Lsun} & 0.546 & 0.568 & 0.507 & \multicolumn{1}{l|}{0.439} & 0.117* & 0.035 & 0.037 & 0.117 \\
\multicolumn{1}{l|}{Target} & 0.629* & 0.587 & 0.420 & \multicolumn{1}{l|}{0.295} & 0.550* & 0.025 & 0.015 & 0.253 \\
\multicolumn{1}{l|}{Tetra} & 0.504* & 0.504* & 0.504* & \multicolumn{1}{l|}{0.504} & 0.200* & 0.200* & 0.200* & 0.200* \\
\multicolumn{1}{l|}{TwoDiamonds} & 0.486* & 0.486* & 0.486* & \multicolumn{1}{l|}{0.486} & 0.044* & 0.022 & 0.022 & 0.022 \\
\multicolumn{1}{l|}{WingNut} & 0.384 & 0.423* & 0.384 & \multicolumn{1}{l|}{0.384} & 0.063* & 0.024 & 0.063* & 0.063*
\end{tabular}
\caption{Comparison of Methods on FCPS Datasets}\label{tab:results}
\end{table}

ICOT dominates the OCT results in all cases for both metrics; this demonstrates the advantage of building clusters directly through a tree-based approach rather than applying a tree to cluster labels \textit{a posteriori}. ICOT matches or outperforms K-Means in 4/9 cases with Silhouette and 8/9 cases with Dunn. We are unable to capture the ground truth when the underlying clusters are nonseparable with parallel splits (i.e. Atom, Hepta datasets). ICOT places hard constraints on an observation's cluster membership based on splits in feature values, whereas K-Means is less constrained. However, this trade-off allows for clear cluster definitions; thus we accept a slight decrease in cluster quality score for the gain in interpretability due to the importance of intuitive assignment rules in many settings.

Cluster quality evaluation is highly dependent on the chosen metric; the ground truth scores lower than the ICOT clusters in 5 of 9 cases for Silhouette and 2 of 9 cases for Dunn. This raises the broader question of how to assess cluster quality; recovering known labels in synthetic data does not necessarily translate to meaningful cluster assignment. The ICOT validation criterion should be chosen in consideration of the desired cluster properties. The Dunn Index performs well at identifying clusters by geometric separation, while the Silhouette Metric is often better at finding meaningful separation when accounting for the density of the data.

\subsection{A Real World Example}

\begin{figure}[h]
\centering
  \includegraphics[width=0.65\linewidth]{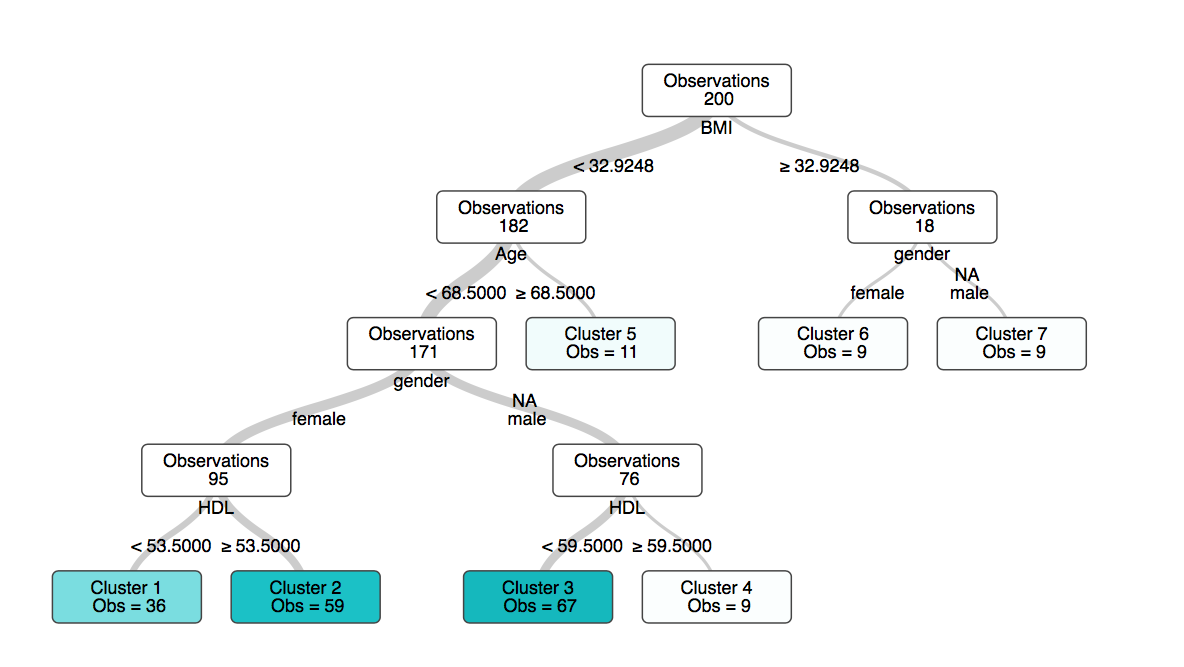}
  \caption{Visualization of the ICOT results for the Dunn index on Framingham Heart Study dataset.}
  \label{fig:FHS_example}
\end{figure}

We provide an illustration of our method using data from the Offspring Cohort from the Framingham Heart Study, a large-scale longitudinal clinical study. The dataset comprises of 200 observations and 8 covariates (age, gender, presence of diabetes, levels of HDL, Systolic Blood Pressure, BMI and hematocrit). The ICOT algorithm creates 7 clusters corresponding to the leaves of the tree in Figure~\ref{fig:FHS_example} and selects only four features to split on. The interpretable nature of ICOT allows us to understand the differentiating factors in these clusters. For example, we see a separation between younger women and men, and furthermore a different HDL threshold to distinguish among participants within each gender. Thus, we are able to clearly define the characteristics of each cluster.

\section{Conclusions}
We have introduced a new methodology of cluster creation that addresses the issue of cluster interpretability. Our method extends the framework of Optimal Classification Trees to an unsupervised learning setting, in which we build trees that provide explicit separations of the data on the original feature set. This makes it an ideal tool for exploratory data analysis since it reveals natural separations of the data with intuitive reasoning. We believe that our proposed clustering algorithm offers a promising alternative to existing methods, namely K-Means and hierarchical clustering. Our early results suggest that we can recover clusters similar to K-Means, but with the added advantages of interpretability and no prespecified cluster count. We hope to apply this method across various applications, particularly in healthcare, including grouping together similar patients, medical diagnoses, and others.

\newpage
\printbibliography
\newpage
\section{Supplementary Material}

The OCT algorithm formulates tree construction as a MIO which allows us to define a single problem, as opposed to the traditional recursive, top-down methods that must consider each of the tree decisions in isolation. It allows us to consider the full impact of the decisions being made at the top of the tree, rather than simply making a series of locally optimal decisions, avoiding the need for pruning and impurity measures.

The MIO framework of Optimal Classification Trees (OCT) can be modified to address an unsupervised learning setting. We present changes in the original MIO formulation of OCT to be able to partition the data space into distinct clusters following the same structure and notation as in \cite{bertsimas2017optimal}. There are two primary modifications in our model:
\begin{enumerate}
    \item The objective function is comprised solely by the chosen cluster quality criterion, such as the Silhouette Metric, and does not include any penalty for the tree complexity.
    \item Each leaf of the tree is equivalent to a cluster. Observations in different leaves are not allowed to belong to the same cluster.
\end{enumerate}
Given a tree object, we will index its nodes by $t=1,\dots,T$. We use the notation $p(t)$ to refer to the parent node of node $t$, and $A(t)$ to denote the set of ancestors of node $t$. We also define $A_L(t)$ as the set of ancestors of $t$ whose left branch has been followed on the path from the root node to $t$, and similarly $A_R(t)$ is the set of the right-branch ancestors.

The nodes in the tree are divided into two sets:
\begin{itemize}
    \item Branch nodes: Nodes $t \in \cal{T_B}$  apply a split of the form $a^\intercal x < b$. All the points that satisfy the split follow the left branch in the tree and those that do not follow the right branch.
    \item Leaf nodes: Nodes $t \in \cal{T_L}$ formulate a cluster for all the points that fall into the leaf node.
\end{itemize}

As in the OCT formulation, we define the split applied at node $t \in \cal{T_B}$ with variables $\mathbf{a}_t \in \mathbb{R}^p$ and $b_t \in \mathbb{R}$. The vector $\mathbf{a}_t$ indicates which variable is chosen for the split, meaning that $a_{jt} = 1$ for the variable $j$ used at node $t$. $b_t$ gives the threshold for the split, which is between $\left[0,1\right]$ after normalization of the feature vector. Together, these form the constraint $\mathbf{a}_t^T x < b_t$. The indicator variables $d_t$ are set to 1 for branch nodes and 0 for leaf nodes. Using the above variables, we introduce the following constraints that allows us to model the tree structure (for a detailed analysis of the constraints, see \cite{bertsimas2017optimal}):
    \begin{align}
       \sum_{j=1}^p a_{jt} = d_t,\; \forall t \in \cal{T_B}, \\
       0 \leq b_t \leq d_t, \; \forall t \in \cal{T_B}, \\
       a_{jt} \in \{0,1\}, \; j=1,\dots,p, \quad \forall t \in \mathcal{T_B},\\
       d_t \leq d_{p(t)}, \; \forall t \in \mathcal{T_B} \backslash \{1\} ,
    \end{align}

Next we present the corresponding constraints that track the allocation of points to leaves. For this purpose, we introduce the indicator variables $z_{it} = \mathbbm{1}$\{$x_i$ is in node $t$\} and $l_t = \mathbbm{1}$\{leaf $t$ contains any points\}. We let $N_{min}$ be a constant that defines the minimum number of observations required in each leaf. We apply the following constraints as in OCT:
\begin{align}
       \sum_{t \in \cal{T_L}} z_{it}=1, \; i=1,\dots,n,\\
       z_{it} \leq l_t, \; \forall t \in \mathcal{T_L},\\
       \sum_{i=1}^n z_{it} \geq N_{min}l_t, \; \forall t \in \mathcal{T_L}
\end{align}

Next we present the set of constraints that enforce the splits that are required by the structure of the tree when assigning points to leaves. We want to enforce a strict inequality for points going to the lower leaf. To accomplish this, we define the vector $\epsilon \in \mathbb{R}^p$ as the smallest separation between two observations in each dimension $p$, and $\epsilon_{max}$ as the maximum over this vector. The split can then be enforced using the following constraints:
\begin{align}
       a_m^\intercal x_i \geq b_t - (1-z_{it}),  \; i = 1,\dots,n, \quad \forall t \in \mathcal{T_B}, \quad \forall m \in A_{R}(t) \\
       a_m^\intercal (x_i + \epsilon) \leq b_t +(1 + \epsilon_{max}) (1-z_{it}),  \; i = 1,\dots,n, \quad \forall t \in \mathcal{T_B}, \quad \forall m \in A_{L}(t) \\
\end{align}

The objective of the new formulation is to maximize the Silhouette Metric $S$ of the overall partition. The Silhouette Metric quantifies the difference in separation between a point and points in its cluster, vs. the separation between that point and points in the second closest cluster.

Let $d_{ij}$ be the distance (i.e. Euclidean) of observation $i$ from observation $j$. We define $K_t$ to be number of points assigned assigned to cluster $t$.
\begin{equation}
    K_t = \sum_{i=1}^n z_{it} \forall t \in \cal{T_L}
\end{equation}
We define $c_it$ to be the average distance of observation $i$ from cluster $t$:
\begin{equation}
    c_{it} = \frac{1}{K_t} \sum_{j=1}^n d_{ij}z_{jt}, \; i = 1, \ldots,  n, \; \forall t \in \cal{T_L}.
\end{equation}

We define $r_i$ to be the average distance of observation $i$ from all the points assigned in the same cluster:
\begin{equation}
    r_i = \sum_{\forall t \in \cal{T_L}} c_{it}z_{it}, \; i = 1, \ldots,  n.
\end{equation}

We then let $w_{it}$ denote the minimum average distance of observation $i$ from all the points in cluster $t$ where $i$ does not belong to cluster $t$.
\begin{equation}
    q_i \leq c_{it}(1-z_{it}) + M z_{it}, \; i = 1, \ldots,  n, \;  \forall t \in \cal{T_L}.
\end{equation}

Finally, to define the Silhouette Metric of observation $i$, we will need the maximum value between $r_i$ and $q_i$ which normalizes the metric.
\begin{equation}
    m_i \geq r_i, \; i = 1, \ldots,  n.
\end{equation}
\begin{equation}
    m_i \geq q_i, \; i = 1, \ldots,  n.
\end{equation}
The Silhouette Metric for each observation is computed as $s(i)$ and the overall Silhouette Metric for the clustering assignment is then the average over all Silhouette Metrics from the training population:
\begin{equation}
    s_i = \frac{q_i - r_i}{m_i}, \; i = 1, \ldots,  n.
\end{equation}
\begin{equation}
    S = \frac{1}{n}\sum_{i=1}^n s_i.
\end{equation}
\end{document}